\documentclass[letterpaper, 10 pt, conference]{ieeeconf}  

\IEEEoverridecommandlockouts                              

\usepackage{graphics} 
\usepackage{amsmath, amsfonts} 
\usepackage{mathrsfs}
\usepackage{amssymb}  
\usepackage{algorithm}
\usepackage[noend]{algpseudocode}
\usepackage{multirow}
\usepackage{array}
\usepackage{booktabs}
\usepackage{threeparttable}
\usepackage{cite}
\usepackage[caption=false,font=footnotesize,labelfont=rm,textfont=rm,subrefformat=parens]{subfig}
\usepackage{stfloats}
\usepackage{cases}
\usepackage{graphicx}

\title{\LARGE \bf
Uncertainty-Aware Safety-Critical Decision and Control for Autonomous Vehicles at Unsignalized Intersections}

\author{Ran Yu, Zhuoren Li, Lu Xiong, Wei Han, Bo Leng
\thanks{This work has been submitted to the IEEE for possible publication.
 Copyright may be transferred without notice, after which this version may
 no longer be accessible. \textit{(corresponding author: Zhuoren Li).}}
\thanks{Ran Yu, Zhuoren Li, Lu Xiong, Wei Han and Bo Leng are with the School of Automotive Studies, Tongji University, Shanghai 201804, China (email:ranyu@tongji.edu.cn; 1911055@tongji.edu.cn; xiong\_lu@tongji.edu.cn; tjhanwei@tongji.edu.cn; lengbo@tongji.edu.cn;).}
}

\begin{document}

\maketitle
\thispagestyle{empty}
\pagestyle{empty}

\begin{abstract}
Reinforcement learning (RL) has demonstrated potential in autonomous driving (AD) decision tasks. However, applying RL to urban AD, particularly in intersection scenarios, still faces significant challenges. The lack of safety constraints makes RL vulnerable to risks. Additionally, cognitive limitations and environmental randomness can lead to unreliable decisions in safety-critical scenarios. Therefore, it is essential to quantify confidence in RL decisions to improve safety. This paper proposes an Uncertainty-aware Safety-Critical Decision and Control (USDC) framework, which generates a risk-averse policy by constructing a risk-aware ensemble distributional RL, while estimating uncertainty to quantify the policy's reliability. Subsequently, a high-order control barrier function (HOCBF) is employed as a safety filter to minimize intervention policy while dynamically enhancing constraints based on uncertainty. The ensemble critics evaluate both HOCBF and RL policies, embedding uncertainty to achieve dynamic switching between safe and flexible strategies, thereby balancing safety and efficiency. Simulation tests on unsignalized intersections in multiple tasks indicate  that USDC can improve safety while maintaining traffic efficiency compared to baselines.
\end{abstract}

\section{Introduction}

Autonomous driving (AD) is becoming a significant focus of global innovation due to its potential in energy saving, emission reduction, traffic efficiency, road safety, and releasing drivers \cite{WANG202417}. Although rule-based autonomous vehicles (AV) are effective in reducing accident rates in straightforward, merge, and exit scenarios, the accident frequency in turning scenarios is 1.98 times higher than that of human-driven vehicles (HDVs) \cite{abdel2024matched}. Due to the limitations of predefined rules in covering all driving scenarios, AVs struggle to make flexible decisions when faced with multiple oncoming HDVs, potentially causing traffic congestion or rear-end collisions \cite{8370800, 20214811249569}.

Recently, reinforcement learning (RL) has demonstrated remarkable proficiency in decision-making tasks, such as those on the highway \cite{10422331}, merge \cite{20233614676836}, and intersection \cite{20244717409481}, due to its ability to optimize high-dimensional state spaces through continuous interaction with dynamic environments. However, even well-trained RL agents encounter considerable challenges in ensuring the safety of policies due to the lack of safety constraints \cite{10740674}. Consequently, Safe RL has emerged as a paradigm for RL applications, with the objective of maximizing the expected cumulative reward while simultaneously satisfying safety constraints \cite{10675394}.

A typical Safe RL architecture is based on the Constrianed Markov Decision Process (CMDP) framework, which guarantees safety by constraining the cumulative expected cost below a threshold. This is usually addressed by solving the constrained optimization problem using Lagrangian multiplier methods \cite{stooke2020responsive, honari2024meta} or trust region methods \cite{achiam2017constrained, zhang2020first}. However, as the policy gradually converges near the safety threshold, learning a safe policy becomes more challenging due to the sparsity of the cost signal. Another common paradigm for Safe RL involves performing safety corrections to the RL policy, such as using action masks \cite{shixin2024unmanned}
and safe energy functions \cite{10402567, 10610959, 9718195}. Action masks identify unsafe actions by predefining a safe action set and limiting unsafe actions from participating in the network's updates. Nevertheless, such methods are typically applicable only to discrete action spaces. 
Control barrier functions (CBF) are a significant representative of safety energy functions and are widely regarded as a well-established method in safety control \cite{8796030}. However, many CBF approaches require the design of intricate prior functions, which limits their adaptability and generalization in complex dynamic scenarios. Furthermore, the existence of high relative degree between the standard CBF and the underlying system control inputs requires a more conservative barrier function for safety, otherwise the safety can be compromised due to the inability to constrain the higher-order dynamics resulting in transient transgressions of the lower-order constraints.  Learning-based safe energy functions have garnered significant attention because of their independence from prior knowledge. However, they are deficient in clearly defined and interpretable safety guarantee mechanisms.

RL decisions typically yield only outcomes without revealing the uncertainty. This inherent ambiguity can pose safety risks in AD tasks \cite{10104197}. This is particularly evident in scenarios involving long-tailed problems or out-of-distribution (OOD) events, where the reliability of RL policies is significantly compromised. The quantification of the confidence level of RL decisions facilitates the identification of RL actions that pose potential safety risks due to high uncertainty. This, in turn, enables the reduction or avoidance of dangerous decisions in OOD scenarios. Uncertainty can be categorized into two main types: epistemic uncertainty (EU) and aleatoric uncertainty (AU) \cite{hullermeier2021aleatoric}. EU reflects data-dependent model or parameter uncertainty, which arises from lack of knowledge and can be reduced by observing more data; AU is irreducible, reflecting the stochastic nature of environmental dynamics \cite{10836828}. Recent works in the field of AD have demonstrated that uncertainty-aware RL has garnered significant attention and focus. A common approach is to use a backup policy to replace the RL policy when the estimated uncertainty exceeds a threshold or falls outside a safety set \cite{10073955, 10534899, 10740674, 10107652, 10155311}. However, the set or threshold is often limited to specific scenarios. Furthermore, some backup strategies, such as deceleration or braking, may negatively impact traffic efficiency.

\begin{figure*}[!t]
\centering
\includegraphics[width=16.0cm]{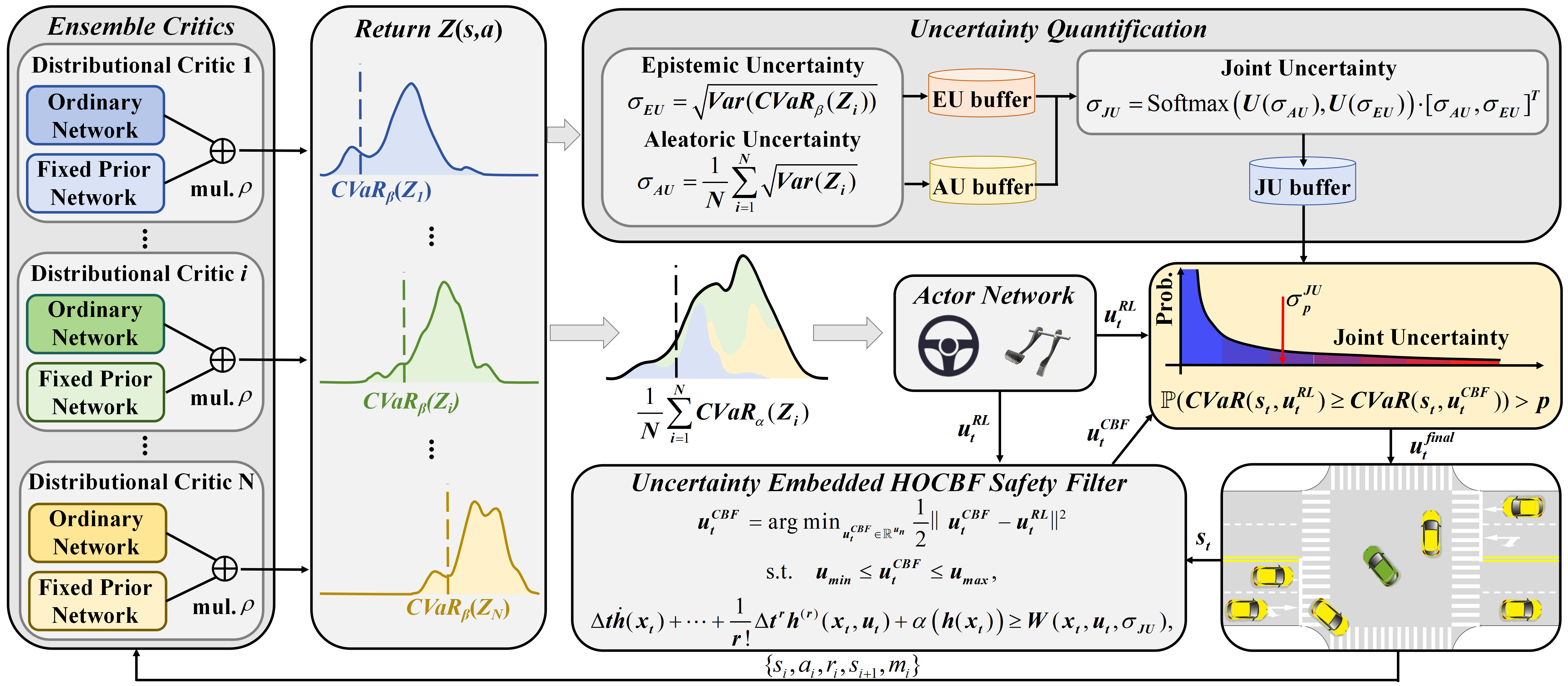}
\caption{Diagram of our framework.}
\label{framework}
\end{figure*}

To this end, in order to enhance the ability of RL to handle safety-critical and OOD scenarios while achieving a balance between safety and traffic efficiency, we propose the Uncertainty-aware Safety-critical Decision and Control (USDC) framework, as illustrated in Fig. \ref{framework}. This framework leverages deep ensemble (DE) \cite{ganaie2022ensemble} to estimate the joint uncertainty (JU) that encompasses both EU and AU. By dynamically adjusting safety constraints based on JU, it effectively balances safety and efficiency. The main contributions are as follows:
\begin{itemize}
\item We propose a risk-aware distributional RL ensemble architecture that combines the original and Fixed Prior Networks (FPN) to build the critic. It quantifies tail risk in the reward distribution and generates a risk-averse policy, while jointly estimating EU and AU for dual uncertainty-driven decision-making.
\item A high-order control barrier function (HOCBF) is constructed, which ensures safety using only relative distance. Moreover, HOCBF incorporates JU to dynamically adjust safety constraints, balancing safety and traffic efficiency.
\item Based on the JU distribution, the HOCBF and RL policies are evaluated, ensuring that the RL policy performs no worse than the HOCBF under low uncertainty and favors safer policies under high uncertainty.
\end{itemize}

\section{Preliminaries}

\subsection{Distributional Reinforcement Learning}
RL can be modeled as a Markov Decision Process (MDP) defined by a tuple $(\mathcal S,\mathcal A,\mathcal P, \mathcal R,\rho_0,\gamma)$.
$\mathcal S$ is the state space, $\mathcal A$ is the action space, $P:\mathcal{S}\times\mathcal{A}\times\mathcal{S}\to[0,1]$ is the transition probability function, $\mathcal{R}:\mathcal{S}\times\mathcal{A}\to\mathbb{R}$ is the reward function, $\rho_0:\mathcal{S}\to[0,1]$ is the initial state distribution, and $\gamma$ is the discount factor for future reward. The policy $\pi:\mathcal{S}\to{\mathcal{P}}(\mathcal{A})$ is a map from given states to a probability distribution on action space. Standard MDP aims at maximizing the agent's cumulative discounted reward $\mathcal{L}(\pi)=\mathbb{E}_{\pi,\mathcal{P}}\left[\sum_{t=0}^\infty\gamma^t r(s_t,a_t)\right]$. For $\pi\in\Pi$, the $Q$ function $Q^\pi:\mathcal{S}\times\mathcal{A}\to\mathbb{R}$ can be defined as $Q^\pi(s,a):=\mathbb{E}_{\pi,\mathcal{P}}\left[\sum_{t=0}^\infty\gamma^t r(s_t,a_t)\right]$, which satisfies the following Bellman equation:
\begin{equation}
\begin{aligned}[b]
 Q(s,a) := \mathbb{E}[r(s,a)] + \gamma \mathbb{E}_{\pi,\mathcal{P}}[Q(s',a')].
\label{E1}
\end{aligned}
\end{equation}

In the distributional RL setup, the distribution over returns $Z(s,a)$ is estimated, and its expectation is maximized to obtain $Q(s,a) = \mathbb{E}[Z(s,a)]$. Hence, we can rewrite (\ref{E1}) as:
\begin{equation}
\begin{aligned}[b]
 Z(s,a) \overset{D}{:=} r(s,a) + \gamma Z(s',a'),
\label{E2}
\end{aligned}
\end{equation} where $\overset{D}{:=}$ indicates that both sides of the equation are distributed according to the same distribution. Let $F_z(z)=\mathbb{P}(Z\leq z)$ denote the cumulative distribution function (CDF) of random variable $Z$, the quantile function $F^{-1}_{Z}$ can be represented as the inverse of CDF. Given quantile fraction $\tau$, we have $F^{-1}_{Z}:=\inf\{z\in\mathbb{R}:\tau\leq F_Z(z)\}$. 

Following \cite{ma2020dsac}, we defined a serious of quantile fractions $\{\tau_i\}_{i=0,...,N}$, $\hat{\tau_i}=(\tau_i + \tau_{i+1}/2)$, and the pairwise temporal difference (TD) error between two quantile fractions $\hat{\tau_i}, \hat{\tau_j}$:
\begin{equation}
\begin{aligned}[b]
\delta_{ij} = r + \gamma \left[ Z^{\bar\theta}_{\hat{\tau_i}}(s', a') - \alpha \log \pi_{\bar{\phi}}(a'|s') \right] - Z^{\theta}_{\hat{\tau_j}}(s, a)
\label{E3}
\end{aligned}
\end{equation} where $\bar{\theta},\theta, \bar{\phi}$ are parameters of target critic, critic, and target actor network, respectively. $\alpha$ is the temperature parameter. We then tarin the critic $Z^{\theta}_{\tau}(s, a)$ using quantile regression by minimizing the weighted pairwise Huber loss \cite{huber1992robust} with threshold $\kappa$:
\begin{equation}
\begin{aligned}[b]
\mathcal{L}_Z(\theta)=\frac{1}{|\mathcal{B}|}\sum_{(s,a,r,s')\in\mathcal{B}}\sum_{i=0}^{N_q-1}\sum_{j=0}^{N_q-1}(\tau_{i+1}-\tau_i)\rho_{\hat{\tau}_j}^\kappa\left(\delta_{ij}\right)
\label{E4}
\end{aligned}
\end{equation}
\vspace{-10pt}
\begin{equation}
\begin{aligned}[b]
\rho_{\tau}^{\kappa}\left(\delta_{ij}\right)=\left|\tau-\mathbb{I}\left\{\delta_{ij}<0\right\}\right|\frac{\mathcal{L}_{\kappa}\left(\delta_{ij}\right)}{\kappa},
\label{E5}
\end{aligned}
\end{equation}
\vspace{-10pt}
\begin{equation}
\begin{aligned}[b]
\left.{\mathcal{L}}_{\kappa}\left(\delta_{ij}\right)=\left\{
\begin{array}
{ll}{\frac{1}{2}\delta_{ij}^{2},} & {\mathrm{~if~}|\delta_{ij}|\leq\kappa} \\
{\kappa\left(|\delta_{ij}|-\frac{1}{2}\kappa\right),} & {\mathrm{~otherwise},}
\end{array}\right.\right.
\label{E6}
\end{aligned}
\end{equation} where $\mathcal{B}$ is a minibatch of transitions sampled from a replay buffer, $N_q$ denotes the number of quantile points over $[0,1]$. Then, the objective of actor is to maximize the $Q$-return:
\begin{equation}
\begin{aligned}[b]
L_\pi(\phi)=\mathbb{E}_{s_t\sim \mathcal{D},a_t\sim\pi_\phi}[\alpha\log(\pi_\phi(a_t|s_t))-Q_\theta(s_t,a_t)],
\label{E7}
\end{aligned}
\end{equation} where $Q_\theta(s,a)=\sum^{N_q}_{i=0}(\tau_{i+1}-\tau_{i})Z^{\theta}_{\hat{\tau_i}}(s,a)$, $\mathcal{D}$ is the transitions replay buffer.

\subsection{High-Order Control Barrier Functions}
Consider an input-affine control system given by:
\begin{equation}
\begin{aligned}[b]
\dot {\boldsymbol{x}} = f(\boldsymbol{x}) + g(\boldsymbol{x})\boldsymbol{u},
\label{E8}
\end{aligned}
\end{equation} where the system state $\boldsymbol{x}\in\mathcal{X}\subset\mathbb{R}^n$ and the control input $\boldsymbol{u}\in\mathcal{U}\subset\mathbb{R}^{u_n}$. $f$ and $g$ are locally Lipschitz.
\newtheorem{definition}{Definition}
\begin{definition} [Forward invariant set]
The set $\mathcal{C}$ is forward invariant for system (\ref{E8}) if for every initial condition $\boldsymbol{x}(t_0)\in \mathcal{C}$, $\boldsymbol{x}(t)\in \mathcal{C}$ for $\forall t\geq t_0$. For a continuously differentiable function $h:\mathcal{X}\rightarrow \mathbb{R}$, let
\begin{equation}
\begin{aligned}[b]
\mathcal{C}=\{\boldsymbol{x}\in\mathcal{X}:h(\boldsymbol{x})\geq0\}.
\label{E9}
\end{aligned}
\end{equation}
\end{definition}

\begin{definition} [CBF \cite{8796030}] 
Denote class $\mathcal{K}$ function is a function $\alpha:[0,a)\to[0,\infty),a>0$ that is strictly increasing and with $\alpha(0)=0$. Given the superlevel set $\mathcal{C}$ as in (\ref{E9}), $h$ is a CBF if there exists a class $\mathcal{K}$ function $\alpha$ such that
\begin{equation}
\begin{aligned}[b]
\sup_{\boldsymbol{u}\in\mathcal{U}}\left[L_{f}h(\boldsymbol{x})+L_{g}h(\boldsymbol{x})\boldsymbol{u}\right]\geq-\alpha(h(\boldsymbol{x})), \forall \boldsymbol{x}\in\mathcal{C},
\label{E10}
\end{aligned}
\end{equation}
\end{definition} where $L_{f}h,L_{g}h$ denote the Lie derivatives along $f$ and $g$, respectively. In the discrete system perspective, we consider a discrete-time system $\boldsymbol{x}_{k+1}=f(\boldsymbol{x}_k)+g(\boldsymbol{x}_k)\boldsymbol{u}_k$, where $k\in\mathbb{N}$ denotes the time step. Then, given a set $\mathcal{C}$ as in (\ref{E9}), continuous function $h$ is a candidate discrete CBF if there exists a class $\mathcal{K}$ function satisfying $\alpha(x)\leq x$ such that:
\begin{equation}
\begin{aligned}[b]
\sup_{\boldsymbol{u}\in\mathcal{U}}\left[\Delta h(\boldsymbol{x},\boldsymbol{u})+\alpha(h(\boldsymbol{x}))\right]\geq0, \forall \boldsymbol{x}\in\mathcal{C},
\label{E11}
\end{aligned}
\end{equation} where $\Delta h(\boldsymbol{x}_k,\boldsymbol{u}_k)=h(\boldsymbol{x}_{k+1})-h(\boldsymbol{x}_{k})$. Note that linear class $\mathcal{K}$ function is commonly used in the discrete domain, i.e., $\alpha(h(\boldsymbol{x}))=\lambda h(\boldsymbol{x}), \lambda\in(0,1]$.

\begin{definition} [Relative degree]
A continuously differentiable function $h:\mathcal{X}\rightarrow\mathbb{R}$ is said to have relative degree $r\in\mathbb{N}$ with respect to system (\ref{E8}), if $\forall x\in\mathcal{X},L_gL_f^ih(\boldsymbol{x})=0, \forall i\in[0,1,...,r-2]$, and $L_gL_f^{r-1}h(\boldsymbol{x})\neq 0$.
\end{definition}

The standard CBF assumes a relative degree of 1, meaning that the first derivative of the safety constraint is equal to the control input. However, this doesn't apply to many autonomous driving constraints, such as the relative degree between distance and acceleration is 2. Therefore, high-order CBFs (HOCBF) are needed to handle such problems.

For the system and the original function $h$ with relative degree $r$, we define the auxiliary function $\Psi_0(x) :=h(\boldsymbol{x})$ and derive the following recursively:
\begin{equation}
\begin{aligned}[b]
\Psi_i(x):=\dot{\Psi}_{i-1}(x)+\alpha_i(\Psi_{i-1}(x)),\forall i\in\{1,\ldots,r\},
\label{E12}
\end{aligned}
\end{equation} where each $\alpha_i$ is a class $\mathcal{K}$ function. Then, a series of superlevel set $\mathcal{C}_i$ can be represented as:
\begin{equation}
\begin{aligned}[b]
\mathcal{C}_i:=\left\{x\in\mathcal{X}:\Psi_{i-1}(x)\geq0\right\},\quad\forall i\in\{1,\ldots,r\}.
\label{E13}
\end{aligned}
\end{equation}
\begin{definition} [HOCBF \cite{9516971, 9777251}]
Given a serious sets $\mathcal{C}_i$ as in (\ref{E13}), a continuously differentiable function $h:\mathcal{X}\rightarrow\mathbb{R}$ is a candidate HOCBF with relative degree $r$ if there exist class $\mathcal{K}$ functions $\alpha_i, \forall i\in\{1,\ldots,r\}$ such that
\begin{equation}
\begin{aligned}[b]
\sup_{\boldsymbol{u} \in \mathcal{U}} \Psi_r(\boldsymbol{x},\boldsymbol{u}) \geq 0,\quad\forall \boldsymbol{x}\bigcap\limits_{i=1}^{r} {\mathcal{C}_i} 
\label{E14}
\end{aligned}
\end{equation}
\end{definition}

\section{Methodologies}
\subsection{Uncertainty-aware Risk-sensitive Distributional RL}
\subsubsection{Risk sensitive RL}
In safety-critical scenarios, we want the AV to adopt conservative measures in the face of high uncertainty, as increased uncertainty typically leads to higher risks. Specifically, when binary safety indicators, such as collision occurrence, are involved, the AV's estimated return distribution tends to follow a multimodal distribution: there are both high return modes corresponding to desirable behaviors and low return modes for unsafe behaviors. To enhance safety, the objective is to minimize the occurrence of low return modes. Thus, we incorporate conditional value-at-risk (CVaR) \cite{ROCKAFELLAR20021443} to formulate a risk-sensitive policy:
\begin{equation}
\begin{aligned}[b]
CVaR_{\beta}(Z)=\mathbb{E}\left[Z|Z<VaR_{\beta}(Z)\right],
\label{E15}
\end{aligned}
\end{equation} where $VaR_{\beta}(Z)$ (the value at risk) is the $\beta$-worst percentile of $Z$. Denote distortion function $\zeta:[0,1]\rightarrow[0,1]$, which is strictly increasing and satisfies $\zeta(0)=0$ and $\zeta(1)=1$. The distorted expectation of $Z$ can be expressed as $\Xi(Z)=\int_0^1F^{-1}_{Z}(\tau)d\zeta(\tau)=\int_{0}^{1}\zeta'(\tau)F^{-1}_{Z}(\tau)d\tau$. And the $CVaR$ is given by $CVaR_{\beta}(Z)=\sum^{N_q-1}_{i=0}(\tau_{i+1}-\tau_{i})\zeta'(\hat{\tau_i})Z^{\theta}_{\hat{\tau}_i}(s,a)$, where $\zeta(\tau)=\min\{\tau/\beta,1\}$.
\subsubsection{Uncertainty Quantification}
We introduce deep ensembles \cite{ganaie2022ensemble} to estimate uncertainty. To improve the recognition ability for OOD scenarios, it is anticipated that the ensemble members demonstrate consistent performance on in-distribution data, whilst preserving diversity on OOD data \cite{rame2021dice}. Specifically, we construct an ensemble critic architecture consisting of $N$ groups of critics $Z^{\theta}_n$ and target critics $Z^{\bar{\theta}}_n$, $n\in[1,..,N]$. Bootstrapping \cite{osband2016deep} was utilized to ensure that each ensemble member has access to a unique subset of the experience replay buffer. Furthermore, we introduce a Randomized Prior Function (RPF) \cite{osband2018randomized}, which adds a fixed prior network (FPN) of the same size as the original network to each ensemble member, improving Bayesian posterior estimates and enhancing diversity on OOD data. The $Z$-return of each ensemble member is composed of two parts:
\begin{equation}
\begin{aligned}[b]
Z_{n,\tau}^{\theta}(s,a)=\frac{\mathcal{O}(s,a;\theta_n) +\rho\mathcal{F}(s,a;\tilde{\theta}_n)}{1+\rho},
\label{E16}
\end{aligned}
\end{equation} where $\mathcal{O}(\cdot), \mathcal{F}(\cdot)$ stand for original network and FPN, respectively. $\tilde{\theta}_n$ are fixed parameters of FPN, $\rho$ is prior factor. Then, the TD-error of $n$-th critic is: 
\begin{equation}
\begin{aligned}[b]
\begin{array}{c}
\delta _{ij}^n = r + \gamma \left[ {\bar Z_{\widehat {{\tau _i}}}^{\bar \theta }(s',a') - \alpha \log {\pi _{\bar \phi }}(a'|s')} \right] - Z_{n,\widehat {{\tau _j}}}^\theta (s,a),
\end{array}
\label{E17}
\end{aligned}
\end{equation} where $\bar Z_{\widehat {{\tau _i}}}^{\bar \theta }(s',a') = \frac{1}{N}\sum_{n = 1}^N {\bar Z_{n,\widehat {{\tau _i}}}^{\bar \theta }(s',a')}$ is the average $Z$-return of $N$ target critic, $a'\sim\pi_{\bar{\phi}}(s')$. The quantile Huber regression loss in (\ref{E4}) is then rewritten as:
\begin{equation}
\begin{aligned}[b]
\mathcal{L}^{n}_Z(\theta)=\frac{1}{|\mathcal{B}|}\sum_{(s,a,r,s')\in\mathcal{B}}\sum_{i=0}^{N_q-1}\sum_{j=0}^{N_q-1}m_{n}(\tau_{i+1}-\tau_i)\rho_{\hat{\tau}_j}^\kappa\left(\delta_{ij}^{n}\right),
\label{E18}
\end{aligned}
\end{equation} where $m_{n}\sim Bernoulli(\mathrm{p}),\mathrm{p}\in(0,1],n\in[1,..N]$ is bootstrap masks sample from Bernoulli distribution. Similarly, the actor objective can be modified as:
\begin{equation}
\begin{aligned}[b]
L_\pi(\phi)=\mathbb{E}_{s_t\sim \mathcal{D},a_t\sim\pi_\phi}[\alpha\log(\pi_\phi(a_t|s_t))-\overline{CVaR}_{\beta}(s,a)],
\label{E19}
\end{aligned}
\end{equation} where $\overline{CVaR}_{\beta}(s,a)=\frac{1}{N}\sum_{n=1}^{N}CVaR_{\beta}(Z^{\theta}_{n})$.

In the context of this paper, epistemic uncertainty is estimated by the disagreement between individual members and the overall ensemble, and is represented through the standard deviation of $CVaR$:
\begin{equation}
\begin{aligned}[b]
{\sigma _{EU}} = \sqrt {Var(CVa{R_\beta }({Z_n^{\theta}}))}. 
\label{E20}
\end{aligned}
\end{equation}

Similar to the approach in \cite{10073955}, we represent aleatoric uncertainty by estimating the standard deviation of the $Z$-return of the ensemble critic:
\begin{equation}
\begin{aligned}[b]
{\sigma _{AU}} = \frac{1}{N}\sum\limits_{n = 1}^N {\sqrt {Var({Z_n^{\theta}})} }.
\label{E21}
\end{aligned}
\end{equation}\

In accordance with the estimated AU and EU, the utilization of joint uncertainty (JU) is employed to account for their combined effect:
\begin{equation}
\begin{aligned}[b]
{\sigma _{JU}} = {\rm{Softmax}}\left( {U({\sigma _{AU}}),U({\sigma _{EU}})} \right) \cdot {[{\sigma _{AU}},{\sigma _{EU}}]^\top},
\label{E22}
\end{aligned}
\end{equation} where $U(x)=1/2(\tanh(x-\mu)/\eta+1)$, $\mu,\eta$ are the mean and variance of $\sigma_{EU}$ or $\sigma_{AU}$, calculated from their buffer, which collects $\mathcal{M}$ steps of $\sigma_{EU}$ and $\sigma_{AU}$. Then, $\sigma _{JU}$ will also be added to the corresponding buffer.

\subsection{Uncertainty-embedded HOCBF}
\subsubsection{Safety Filter Formulation}

\begin{figure}[!t]
\centering
\includegraphics[width=4.5cm]{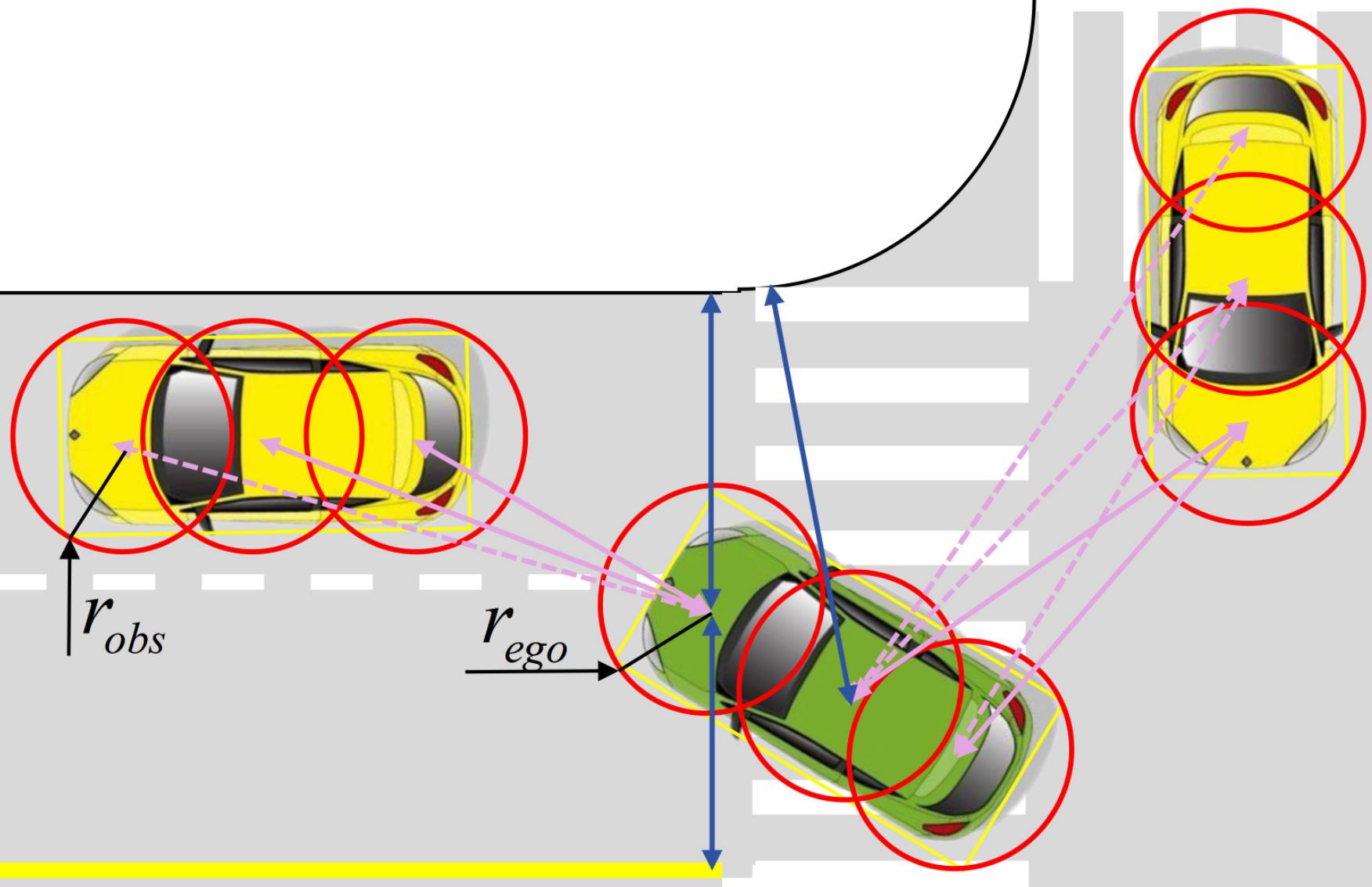}
\caption{Diagram of relative distance. At the \(k\)-th time step, the envelopes of the ego vehicle (EV) and surrounding vehicles (SV) are represented by three circles, denoted as \((x^{\text{ego}}_{i|k}, y^{\text{ego}}_{i|k})\) and \((x^{\text{obs}}_{i|k}, y^{\text{obs}}_{i|k})\), \(i \in \{1, 2, 3\}\). Then, we select \(\mathbf{N}\) closest points to the EV, \((x^{\text{obs}}_{j|k}, y^{\text{obs}}_{j|k})\), \(j \in \{1, ..., \mathbf{N}\}\), and \(\mathbf{M}\) constraints related to the road boundaries. The construction of distance constraints are formulated as a total of \(\{\mathbf{N} \times 3 + \mathbf{M}\}\) constraints.}
\label{dist}
\end{figure}

After obtaining the risk-averse policy $\boldsymbol{u}^{RL}_k$, we introduce HOCBF as a safety filter to further enhance safety. Kinematic bicycle model \cite{rajamani2011vehicle} is utilized to describe the motion of AV:
\begin{equation}
\begin{aligned}[b]
{\boldsymbol{x}_{k + 1}} = \left[ {\begin{array}{*{20}{c}}
{{x_k} + \Delta t \cdot {v_k}\cos ({\varphi _k})}\\
{{y_k} + \Delta t \cdot {v_k}\sin ({\varphi _k})}\\
{{v_k} + \Delta t \cdot {a_{{\rm{lon}}}}}\\
{{\varphi _k} + \Delta t \cdot {v_k}\tan {\delta _f}/L}
\end{array}} \right]
\end{aligned}
\label{E23}
\end{equation} where $\boldsymbol{x}=[x,y,v,\varphi]^\top$, $x,y$ are the position coordinates, $v,\varphi,a_{\rm{lon}},\delta_f,\Delta t, L$ are velocity, heading angle, longitudinal acceleration, steering angle of front wheel, sample time and wheelbase, respectively. Based on the relative distance depicted in Fig. \ref{dist}, we select the candidate CBF as follows:
\begin{equation}
\begin{aligned}[b]
 h_{\text{veh}}(\boldsymbol{x}_k)&=(x_{k}-x^{\text{obs}}_{k})^2+(y_{k}-y^{\text{obs}}_{k})^2 - (r_{obs}+r_{ego})^2,\\
 h_{\text{road}}(\boldsymbol{x}_k)&=(x_{k}-x^{\text{road}}_{k})^2+(y_{k}-y^{\text{road}}_{k})^2 - r_{ego}^2,
\label{E24}
\end{aligned}
\end{equation} where $(x^{\text{road}}_{k},y^{\text{road}}_{k})$ denote the position of road boundary point, and the candidate CBF has relative degree $r=2$.

Consider a discrete-time system as described in (\ref{E23}). To address the problem of high relative degree, we introduce the Truncated Taylor CBF (TTCBF) \cite{xu2025high}, which approximates the discrete-time CBF condition using a truncated Taylor series and requires only a class $\mathcal{K}$ function. For a system with relative degree $r$, TTCBF approximates $\Delta h(\boldsymbol{x}_k,\boldsymbol{u}_k)$ in (\ref{E11}) as $\Delta h(\boldsymbol{x}_k,\boldsymbol{u}_k)\approx \Delta t\dot{h}(\boldsymbol{x}_k)+\frac{1}{2}\Delta t^2\ddot{h}(\boldsymbol{x}_k)+\cdots+\frac{1}{r!}\Delta t^rh^{(r)}(\boldsymbol{x}_k,\boldsymbol{u}_k)$, with the $r$-th derivative $h^{(r)}(\boldsymbol{x}_k,\boldsymbol{u}_k)$ capture the control input.

\begin{definition} [TTCBF \cite{xu2025high}]
Given a set $\mathcal{C}$ as in (\ref{E9}), a continuously differentiable function $h$ is a candidate TTCBF with relative degree $r$ if there exist class $\mathcal{K}$ functions $\alpha(x)\leq x$ such that $\forall \boldsymbol{x}_k \in \mathcal{C}$,
\begin{equation}
\begin{aligned}[b]
\begin{gathered}
\sup_{\boldsymbol{u}_k\in\mathcal{U}}[\Delta t\dot{h}(\boldsymbol{x}_k)+\cdots+\frac{1}{r!}\Delta t^rh^{(r)}(\boldsymbol{x}_k,\boldsymbol{u}_k)+ \\
\alpha\left(h(\boldsymbol{x}_k)\right)]\geq\Gamma\Delta t^{r+1},
\end{gathered}
\end{aligned}
\label{E25}
\end{equation} where $\Gamma$ is a hyper-parameter satisfying $\frac{\Gamma} {(r+1)!}t^{r+1}\geq|R_{r+1}|=\frac{|h^{(r+1)}(\boldsymbol{\xi})|} {(r+1)!}t^{r+1}, \boldsymbol{\xi}\in[\boldsymbol{x}_{k},\boldsymbol{x}_{k+1}]$.
\end{definition}

The first and second time derivatives of $h$ can be derived from (\ref{E23}) and (\ref{E24}) as:
\begin{equation}
\begin{aligned}[b]
& \dot h(\boldsymbol{x}_k)=2(x_{k}-x^{\text{point}}_{k})v\cos\varphi+2(y_{k}-y^{\text{point}}_{k})v\sin\varphi,\\
& \ddot{h}(\boldsymbol{x}_k)=2v^2+2a_{\text{lon}}\Delta_{\text{lon}}+2v^2\tan{\delta_f}\Delta_{\text{lat}}/L,
\end{aligned}
\label{E26}
\end{equation} where $\text{point}\in{[\text{obs},\text{road}]}$, $\Delta_{\text{lon}}=(x_k - x_k^{\text{point}})\sin\varphi + (y_k - y_k^{\text{point}})\cos\varphi$, $\Delta_{\text{lat}}=-(x_k - x_k^{\text{point}})\sin\varphi + (y_k - y_k^{\text{point}})\cos\varphi$. 

Then, substitute $\dot{h}(\boldsymbol{x}_k),\ddot{h}(\boldsymbol{x}_k)$ into (\ref{E25}):
\begin{equation}
\begin{aligned}[b]
\Delta t\dot{h}(\boldsymbol{x}_k)+\frac{1}{2!}\Delta t^2\ddot h(\boldsymbol{x}_k,\boldsymbol{u}_k)+ 
\alpha\left({h(\boldsymbol{x}_k)}\right)\geq\Gamma\Delta t^{3},\\
\Rightarrow \Delta t^2\begin{bmatrix}a_{\text{lon}}\Delta_{\text{lon}},v^2\Delta_{\text{lat}}/L\end{bmatrix}\begin{bmatrix}a_{\text{lon}}\\\tan\delta\end{bmatrix}+h_{\text{cst}}-\Gamma\Delta t^{3}\geq 0,
\end{aligned}
\label{E27}
\end{equation} where $h_{\text{cst}}=\Delta t\dot{h}(\boldsymbol{x}_k)+\alpha\left({h(\boldsymbol{x}_k)}\right)+\Delta t^2v^2$. 

The above inequality (\ref{E27}) is denoted by $\mathcal{H}$.  Similarly, we derive the inequality for all \(\{\mathbf{N}\times3+\mathbf{M}\}\) constraints. The objective of the safety filter is to ensure safety constraints while minimizing modifications to $\boldsymbol{u}^{RL}_k$. Consequently, the following optimization problem can be formulated:
\begin{equation}
\begin{aligned}[b]
\boldsymbol{u}^{CBF}_k  =&\arg\min_{\boldsymbol{u}\in\mathcal{U}}\quad\frac{1}{2}\|\boldsymbol{u}^{CBF}_k-\boldsymbol{u}_k^{RL}\|^2 \\ 
\mathrm{s.t.}& \quad \boldsymbol{u}_{\min}\leq \boldsymbol{u}^{CBF}_k\leq \boldsymbol{u}_{\max},\\
\mathcal{H}_{\text{veh}}=&[\mathcal{H}_{\text{veh}}^{1},..,\mathcal{H}^{\mathbf{N}}_{\text{veh}},\mathcal{H}^{\mathbf{N+1}}_{\text{veh}},...,\mathcal{H}^{\mathbf{2N}}_{\text{veh}},\mathcal{H}^{\mathbf{2N+1}}_{\text{veh}},...,\mathcal{H}^{\mathbf{3N}}_{\text{road}}]^\top\\
\mathcal{H}_{\text{road}}=&[\mathcal{H}_{\text{road}}^{1},..,\mathcal{H}^{\mathbf{M}}_{\text{road}}]^\top,
\end{aligned}
\label{E28}
\end{equation} where $\boldsymbol{u}_{\min},\boldsymbol{u}_{\max}$ represent the input constraints. The optimization problem will be solved using the CasADi nonlinear programming solver \cite{Andersson2018}.

\subsubsection{Uncertainty-embedded Constraints and Probabilistic Improvement}
We expect AV tend to execute safe maneuvers at higher uncertainty, while allowing for greater flexibility under low uncertainty, rather than always executing conservative maneuvers and thus sacrificing passage efficiency. Consider an uncertain system dynamics, $\dot {\boldsymbol{x}} = \hat{f}(\boldsymbol{x}) + \hat{g}(\boldsymbol{x})\boldsymbol{u}+\varphi(\boldsymbol{x},\boldsymbol{u})$, where $\hat{f},\hat{g}$ are known nominal dynamics, $\varphi(\boldsymbol{x},\boldsymbol{u})\in\mathcal{X}\times\mathcal{U}\subset\mathbb{R}^{n}$ is the uncertain vector to be estimated. The uncertainty is then embedded into CBF constraint, such that $\forall \boldsymbol{x}\in\mathcal{C}$:
\begin{equation}
\begin{aligned}[b]
\sup_{\boldsymbol{u}\in\mathcal{U}}\inf_{w\in\mathcal{W}}\left[L_{\hat{f}}h(\boldsymbol{x})+L_{\hat{g}}h(\boldsymbol{x})\boldsymbol{u}+w(\boldsymbol{x},\boldsymbol{u})\right]\geq-\alpha(h(\boldsymbol{x})),
\label{E29}
\end{aligned}
\end{equation} where $w(\boldsymbol{x},\boldsymbol{u})=L_{\varphi}h$ is the uncertain scalar. Then, (\ref{E25}) can be rewritten as:
\begin{equation}
\begin{aligned}[b]
\sup_{\boldsymbol{u}_k\in\mathcal{U}}\inf_{w\in\mathcal{W}}[\Delta t\dot{h}(\boldsymbol{x}_k)+\cdots+\frac{1}{r!}\Delta t^rh^{(r)}(\boldsymbol{x}_k,\boldsymbol{u}_k)+ \\
w(\boldsymbol{x}_k,\boldsymbol{u}_k)+\alpha\left(h(\boldsymbol{x}_k)\right)]\geq\Gamma\Delta t^{r+1}.
\label{E30}
\end{aligned}
\end{equation}

We then make $\Gamma\Delta t^{r+1}-w(\boldsymbol{x}_k,\boldsymbol{u}_k)\rightarrow\mathscr{W}(\boldsymbol{x}_k,\boldsymbol{u}_k,\sigma_{JU})$, which is a positively correlated function of $\sigma_{JU}$. As $\sigma_{JU}$ increases, the constraint becomes stricter.

Furthermore, we calculate the percentile $p$ of the current step's $\sigma_{JU,k}$ within the collected data and utilize ensemble critics to evaluate both $\boldsymbol{u}^{RL}_k$ and $\boldsymbol{u}^{CBF}_k$. $\boldsymbol{u}^{RL}_k$ is selected when the following condition is satisfying:
\begin{equation}
\begin{aligned}[b]
\mathbb{P}(CVaR({\boldsymbol{s}_k},\boldsymbol{u}_k^{RL}) \ge CVaR({\boldsymbol{s}_k},\boldsymbol{u}_k^{CBF})) > p,
\label{E31}
\end{aligned}
\end{equation} where the probability is calculated as the proportion of ensemble critics that consider $\boldsymbol{u}^{RL}_k$ superior to $\boldsymbol{u}^{CBF}_k$, relative to the total number of critics.

\section{Implementations}
\subsection{Environment Settings}
\label{env_set}
We implement a bidirectional four-lane intersection scenario based on Highway-Env \cite{highwayenv}. Each SV is controlled by an improved Intelligent Driver Model (IDM) \cite{idm}, which predicts its heading and position for the subsequent 2 s, yielding to potential collisions according to road priorities. When resetting the scenario, 10 SVs are initialized with random velocities between 6 m/s and 10 m/s. Concurrently, the EV is placed in a collision-free lane with a random velocity. Simulation frequency $f_{s}$ is 15 Hz, with the policy execution frequency $f_{\pi}$ set to 5 Hz during training and 10 Hz during testing.

\subsection{MDPs Design}
\subsubsection{Observation Space and Action Space} The observation space includes the states of the EV $\mathcal{S}_{EV}$ and SVs $\mathcal{S}_{SV}$, the subsequent $N_{wp}$ global reference waypoints $\mathcal{S}_{wp}$, and a one-hot task encoding $\mathcal{S}_{task}$ for left turn, go straight, or right turn: $\mathcal{S}=[\mathcal{S}_{EV},\mathcal{S}_{SV},\mathcal{S}_{wp},\mathcal{S}_{task}]$, where ${S}_{EV}$ = $[\mathbb{I}_{veh}$,$x$,$y$,$v_{x}$,$v_{y}$,$\varphi$,$\omega$,$d_{veh}$,$d_{road}$,$d_{des}]$,  $x,y$ are the position coordinates of the vehicle's center of gravity, $v_x, v_y$ are the longitudinal and lateral velocities, $\varphi$ is the heading angle, $\omega$ is the yaw rate. $\mathbb{I}_{veh} \in \{0,1\}$ is an indicator function to identify whether a vehicle is observed ($\mathbb{I}_{veh}=1$). $d_{veh}, d_{road}, d_{des}$ are the distance to the closest vehicle, the road boundary, and the destination. $\mathcal{S}_{SV}$ contains $N_{SV}$ SV's states: $\mathcal{S}^{j}_{SV}=[\mathbb{I}_{veh}$,$\Delta x$,$\Delta y$,$\Delta v_{x}$,$\Delta v_{y},\varphi]$, $j\in[1,...,N_{SV}]$. $\Delta x,\Delta y,\Delta v_{x},\Delta v_{y}$ represent the position and velocity of the SVs relative to the EV. The waypoints in $\mathcal{S}_{wp}$  are also represented as the relative distances to the EV. The continuous action space is $\mathcal{A}=[a_{\text{lon}}, \delta_f]$.

\subsubsection{Reward Design} 
The reward function consists of sparse \(\mathbf{r}_{sparse}\) and dense \(\mathbf{r}_{dense}\) rewards. Sparse rewards penalize collisions and encourage reaching target points:
\begin{subequations}
\label{E32}
\begin{align}
&\mathbf{r}_{sparse} = \mathbf{r}_{collision} + \mathbf{r}_{arrive\_goal}, \label{E32a} \\
&\mathbf{r}_{collision} = -50 \cdot \mathbb{I}_{collision}, \label{E32b} \\
&\mathbf{r}_{arrive\_goal} = 50 \cdot \mathbb{I}_{arrive\_goal}. \label{E32c}
\end{align}
\end{subequations}

The dense reward considers factors like reference line, action smoothness, distance to destination, and safety distance.
\begin{subequations}
\label{E33}
\begin{align}
&\mathbf{r}_{dense}=3/(1+\mathbf{r}_{ref})+\mathbf{r}_{smooth}+\mathbf{r}_{des}+\mathbf{r}_{safe},
\label{E33_a}\\
&\mathbf{r}_{ref}=\max_{i=1,2}(\boldsymbol{x}_{k,i}^\mathrm{ref}-\boldsymbol{x}_{k})^\top   Q(\boldsymbol{x}_{k,i}^\mathrm{ref}-\boldsymbol{x}_{k}),
\label{E33_b}\\
&\mathbf{r}_{act}=-(\boldsymbol{u}_k^\top R_{\boldsymbol{u}} \boldsymbol{u}_k + R_{\Delta}\Delta \boldsymbol{u}_k), \quad\mathbf{r}_{des}=- d_{des}^2,
\label{E33_c}\\
&\mathbf{r}_{safe}=\begin{cases} 0.0 &  d_{veh} > 5.0, \\ 
-(1.0 - d_{veh}) & 0.5 < d_{veh} \leq 2.0, \\ 
-3 \times (1.0 - d_{veh}) &  d_{veh} \leq 0.5. \end{cases}
\label{E33_d}
\end{align}
\end{subequations} where $\boldsymbol{x}_{k}^\mathrm{ref}= [x^\mathrm{ref},y^\mathrm{ref},v_{x}^\mathrm{ref},0,\phi^\mathrm{ref},0]^\top$. For $\mathbf{r}_{ref}$, both reference lines are computed, and the maximum value is selected to guide the EV to stay close to the reference line. $\mathbf{r}_{act}$ is designed to promote energy efficiency and smooth the trajectory. The weight coefficients are $Q=diag(400.0, 400.0, 30.0, 30.0, 2.0, 0.5)$, $R_{a}=diag(0.05, 0.02)$, $R_{\Delta}=[0.2,0.3]$.

\subsection{Network Architecture and Hyper-Parameters}
\begin{figure}[!t]
\centering
\includegraphics[width=8.5cm]{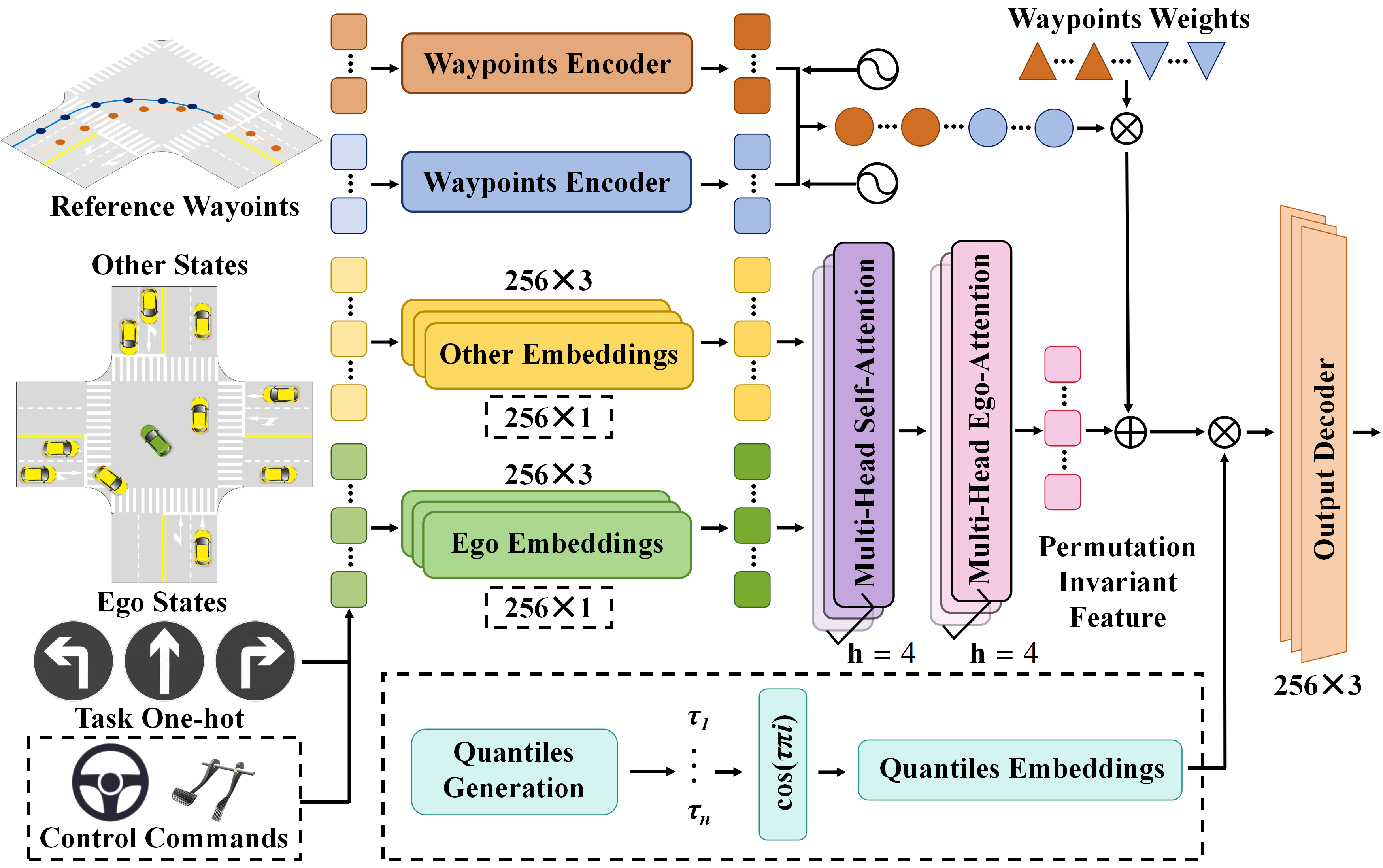}
\caption{Details of designed network. The components within the dashed box are used only in the critic network.}
\label{net}
\end{figure}

As shown in Fig. \ref{net}, the states of SVs, EV, waypoints, control signals, etc. are input into their respective encoding layers. For the waypoints, we apply cosine position encoding and employ learnable weights to capture the varying importance of each waypoint. Based on our previous work \cite{leng2025risk}, we use a 2-hop attention structure to process the information of both SVs and the EV, thereby generating permutation-invariant features. The focus of this paper is not on the method for obtaining quantiles. Consequently, while fixed quantiles are utilized in the experiments, this does not preclude the use of more advanced techniques for quantile generation. 

\subsection{Experiment Results}
We compare USDC with the following baselines: distributional soft actor critic (DSAC \cite{ma2020dsac}); pure-DSAC, which uses a similarly structured multi-layer perception (MLP) network for the network part; USDC without using HOCBF as a safety filter; and Dir-CBF, which directly applies HOCBF filtering to each action generated by a well-trained USDC agent without considering uncertainty. During training, each algorithm is trained three times with different random seeds. In each episode, tasks with randomly generated EVs are assigned, including left turn (LT), go straight (GS), and right turn (GT). During the testing phase, each algorithm is evaluated on 200 episodes for each of the three tasks.

\begin{figure}[!t]
    \centering
    \subfloat{%
        \includegraphics[width=0.23\textwidth]{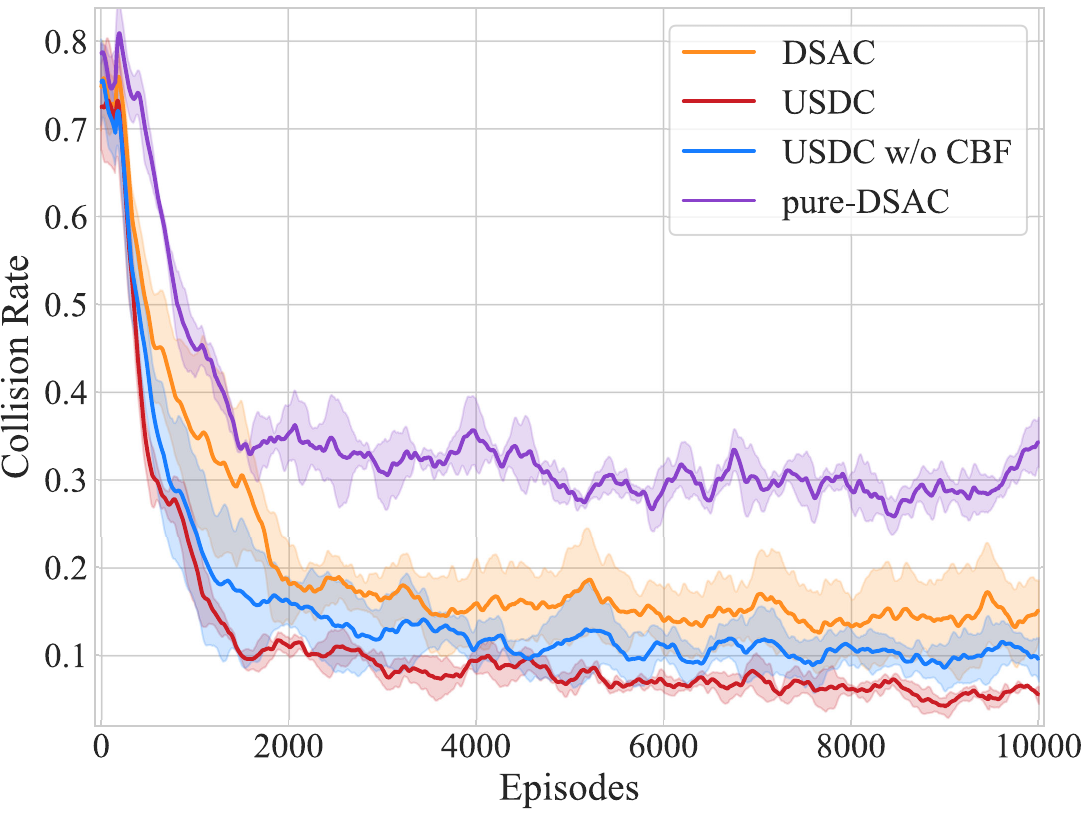}
        \label{c_c}
    }\subfloat{%
        \includegraphics[width=0.23\textwidth]{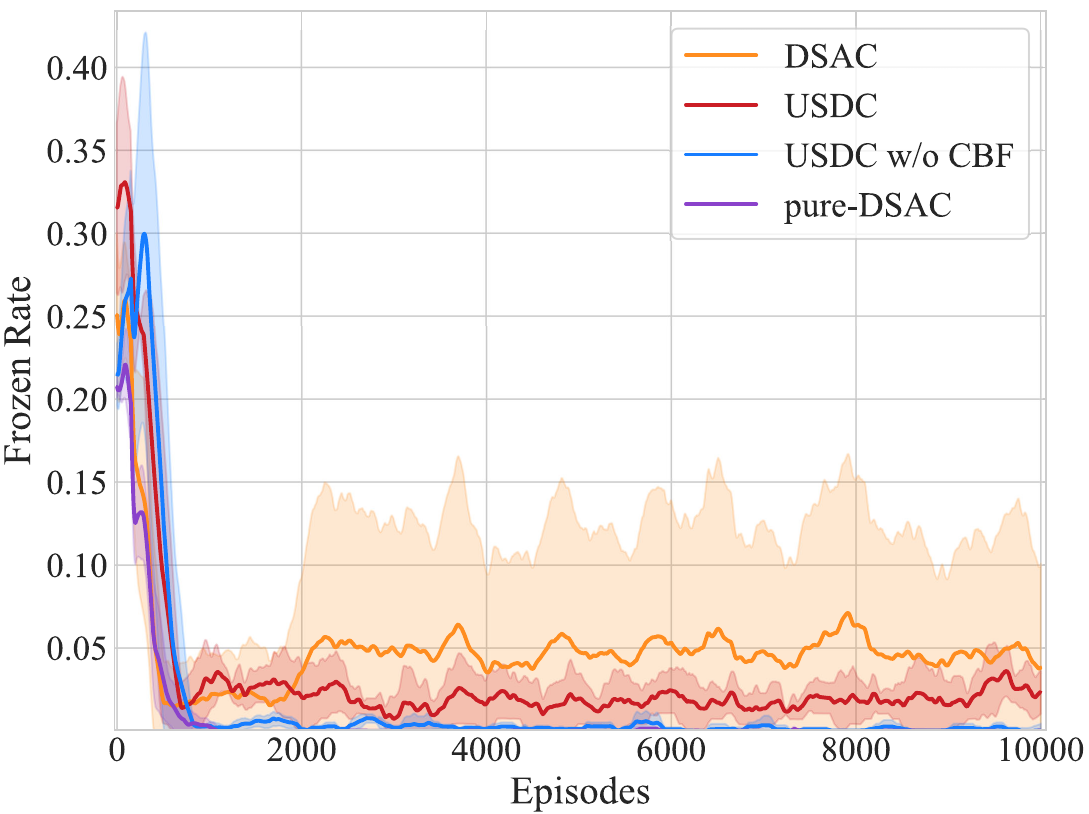}
        \label{c_f}
    }\hfill
    \subfloat{%
        \includegraphics[width=0.23\textwidth]{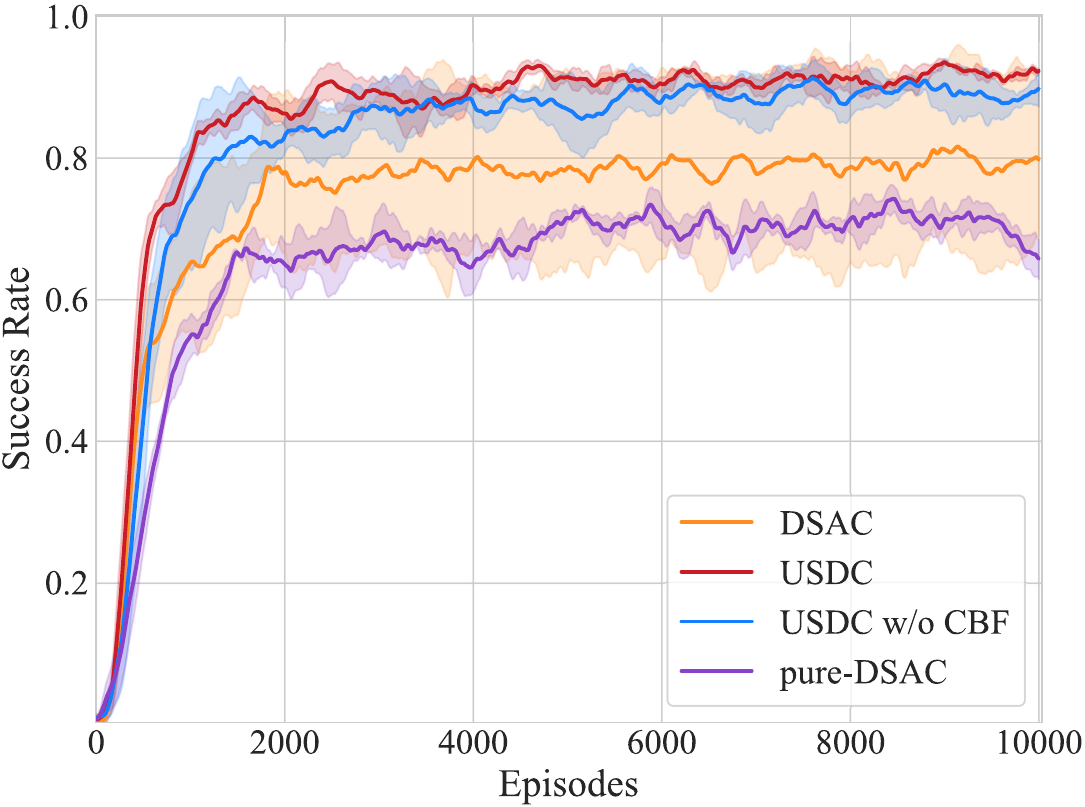}
        \label{c_s}
    }\subfloat{%
        \includegraphics[width=0.23\textwidth]{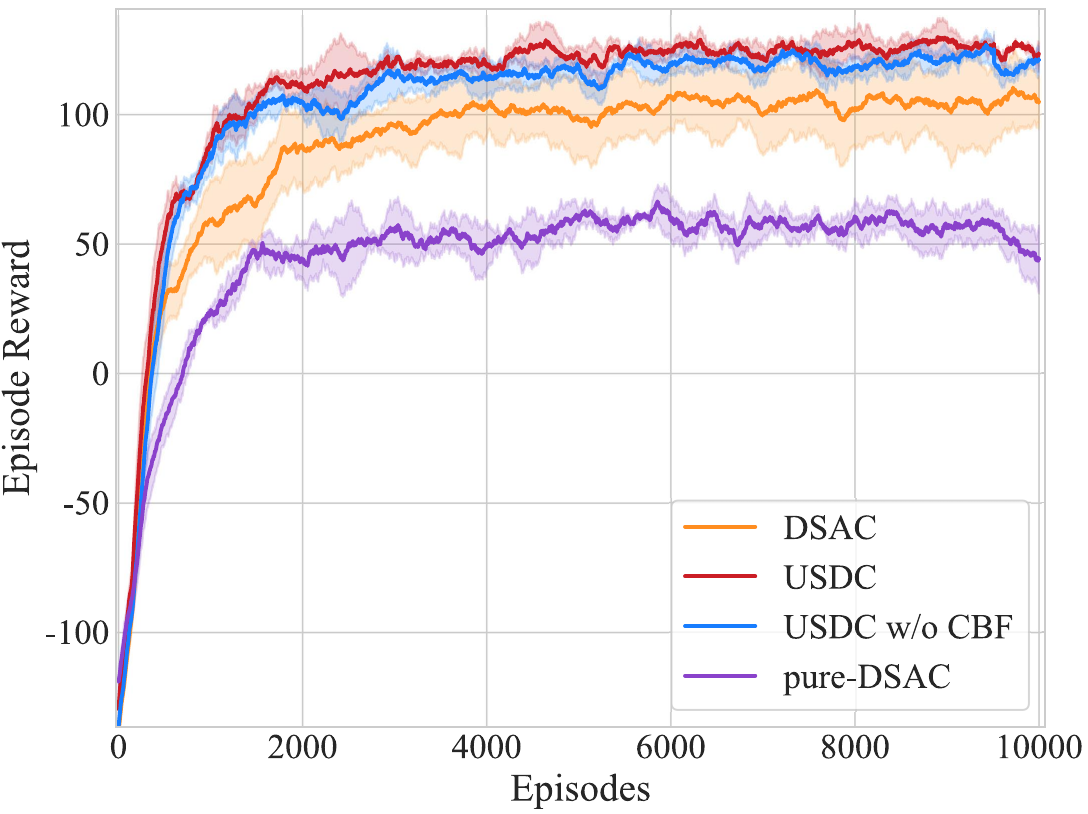}
        \label{c_r}
    }
    \caption{Training curves of experiment. Solid lines correspond to the mean and the shaded regions correspond to 95\% confidence interval over 3 runs.}
    \label{f_compare}
\end{figure}

The learning curves in Fig. \ref{f_compare} and the test results in Table \ref{t_compare} indicate that the USDC with a probabilistic safety filter improves performance while balancing traffic efficiency and safety. Despite an increase in FR compared to USDC without HOCBF, integrating the critic to estimate the $CVaR$ of $\boldsymbol{u}^{RL}$ and $\boldsymbol{u}^{CBF}$ provides safety guarantees under high uncertainty, resulting in a 11.0\% reduction in CR compared to DSAC. In low uncertainty scenarios, the policy that maximizes $CVaR$ is selected by comprehensively assessing the long-term returns and instantaneous risks of $\boldsymbol{u}^{RL}$ and $\boldsymbol{u}^{CBF}$. This results in final performance metrics of 207.3 for AER and 7.77 m/s for AEV, matching or even surpassing other baselines. It is noteworthy that, although USDC increases FR by 3.5\%  compared to the USDC without HOCBF, its dynamic policy switching achieved through uncertainty quantification effectively avoids the drawbacks caused by over-conservatism in direct CBF. As shown in Table \ref{t_compare}, although the direct CBF method controls CR at 2.8\%, its FR is as high as 27.8\%. Additionally, the SR, AER, and AEV are reduced by 19.9\%, 49.3\%, and 12.6\% respectively compared to USDC, indicating that relying solely on safety constraints may compromise the balance between efficiency and safety for AV.

\begin{table}
    \setlength\tabcolsep{3pt}
    \centering
    \caption{Compare Performance on Three Driving Tasks.}
    \begin{threeparttable}
    \scriptsize
    \begin{tabular}{c|c|ccccc}
        \toprule 
        Tasks & Algorithms & SR(\%) 
 & FR(\%) & CR(\%) & AER & AEV(m/s)\\ 
        \midrule 
        \multirow{5}*{LT} & Pure-DSAC & 56.5 & \textbf{0.5} & 43.0 & -22.7$\pm$48.2 & \textbf{7.99$\pm$0.64}\\
        ~ & DSAC & 82.5 & 1.0 & 19.5 & 100.9$\pm$47.5 & 7.00$\pm$1.33\\
        ~ & Dir-CBF & 72.0 & 28.0 & \textbf{4.5} & 6.1$\pm$64.0 & 6.34$\pm$1.51 \\
        ~ & USDC (ours) & \textbf{91.0} & 5.0 & 6.0 & \textbf{129.9$\pm$47.1} & 7.71$\pm$0.44 \\
        ~ & USDC w/o CBF & 86.5 & \textbf{0.5} & 15.5 & 117.4$\pm$46.8 & 7.91$\pm$0.45\\
        \midrule 
        \multirow{5}*{GS} & Pure-DSAC & 62.5 & 3.5 & 34.0 & 30.8$\pm$52.4 & \textbf{7.85$\pm$0.74}\\
        ~ & DSAC & 73.0 & 7.5 & 23.0 & 147.2$\pm$53.4 & 7.73$\pm$0.53\\
        ~ & Dir-CBF & 56.0 & 48.0 & \textbf{3.0} & 58.8$\pm$69.5 &  6.81$\pm$1.04\\
        ~ & USDC (ours) & \textbf{88.5} & 9.0 & 5.0 & \textbf{182.7$\pm$47.7} & 7.59$\pm$0.37\\
        ~ & USDC w/o CBF & 83.5 & \textbf{3.0} & 14.5 & 172.4$\pm$49.5 & 7.57$\pm$0.43\\
        \midrule
        \multirow{5}*{RT} & Pure-DSAC & 88.5 & 0.5 & 11.5 & 190.8$\pm$30.1 & 7.98$\pm$0.57\\
        ~ & DSAC & 98.0 & 0.5 & 1.5 & 267.8$\pm$19.5 & \textbf{8.06$\pm$0.42}\\
        ~ & Dir-CBF & 92.0 & 7.5 & 1.0 & 232.0$\pm$37.1 &  7.21$\pm$0.87\\
        ~ & USDC (ours) & \textbf{100.0} & \textbf{0.0} & \textbf{0.0} & \textbf{309.2$\pm$12.1} & 8.01$\pm$0.22\\
        ~ & USDC w/o CBF & 98.5 & \textbf{0.0} & 1.5 & 278.9$\pm$17.1 & 7.99$\pm$0.29\\
        \midrule 
        \multirow{5}*{MEAN} & Pure-DSAC & 69.2 & 1.5 & 30.2 & 66.3$\pm$43.6 & \textbf{7.94$\pm$0.65}\\
        ~ & DSAC & 84.5 & 3.0 & 14.7 & 172.0$\pm$40.1 & 7.60$\pm$0.76\\
        ~ & Dir-CBF & 73.3 & 27.8 & \textbf{2.8} & 99.0$\pm$56.8 &  6.79$\pm$1.14\\
        ~ & USDC (ours) & \textbf{93.2} & 4.7 & 3.7 & \textbf{207.3$\pm$35.6} &7.77$\pm$0.34\\
        ~ & USDC w/o CBF & 89.5 & \textbf{1.2} & 10.5 & 189.6$\pm$37.8 & 7.82$\pm$0.39\\
        \bottomrule
    \end{tabular}
    \begin{tablenotes}
    \footnotesize
    \item[1] SR, FR, CR, AER, and AEV stand for success rate, frozen rate, collision rate, average episode reward, and average episode velocity, respectively. The definitions are similar to those in \cite{leng2025risk}, except that FR is defined as a runtime exceeding 20 seconds (200 steps).
    \item[2] Bold: best performance. Policy update frequency $f_{\pi} = 10$ Hz.
    \end{tablenotes}
    \end{threeparttable}
    \label{t_compare}
\end{table}

\section{Conclusions}
In this paper, we propose an uncertainty-aware safety-critical decision and control framework. USDC generates a risk-averse policy by constructing a risk-aware distributional ensemble RL, while estimating uncertainty to quantify the reliability of the policy. FPN and Bootstrapping are employed to enhance the diversity of critics when encountering safety-critical scenarios. Subsequently, a HOCBF is employed as a safety filter to minimize intervention policy, while dynamically enhancing constraints based on uncertainty. Instead of relying on a specific threshold, the ensemble critics evaluate both HOCBF and RL policies simultaneously based on the JU distribution. This ensures that the RL policy performs no worse than the HOCBF under low uncertainty and favors safer policies under high uncertainty. Experimental results for left turn, right turn, and go straight driving tasks demonstrate that USDC improves safety compared to baseline methods while maintaining traffic efficiency. The forward invariance of the HOCBF may be violated because of input constraints. Additionally, the HOCBF in this paper only guarantees safety for a single step and does not account for the effects of changes in obstacle states. Future work can be extended to a more robust safety filter to improve overall safety.






\bibliographystyle{IEEEtran}
\bibliography{IEEEabrv,reference}

\begin{thebibliography}{10}
\providecommand{\url}[1]{#1}
\csname url@samestyle\endcsname
\providecommand{\newblock}{\relax}
\providecommand{\bibinfo}[2]{#2}
\providecommand{\BIBentrySTDinterwordspacing}{\spaceskip=0pt\relax}
\providecommand{\BIBentryALTinterwordstretchfactor}{4}
\providecommand{\BIBentryALTinterwordspacing}{\spaceskip=\fontdimen2\font plus
\BIBentryALTinterwordstretchfactor\fontdimen3\font minus \fontdimen4\font\relax}
\providecommand{\BIBforeignlanguage}[2]{{%
\expandafter\ifx\csname l@#1\endcsname\relax
\typeout{** WARNING: IEEEtran.bst: No hyphenation pattern has been}%
\typeout{** loaded for the language `#1'. Using the pattern for}%
\typeout{** the default language instead.}%
\else
\language=\csname l@#1\endcsname
\fi
#2}}
\providecommand{\BIBdecl}{\relax}
\BIBdecl

\bibitem{WANG202417}
H.~Wang, W.~Shao, C.~Sun, K.~Yang, D.~Cao, and J.~Li, ``A survey on an emerging safety challenge for autonomous vehicles: Safety of the intended functionality,'' \emph{Engineering}, vol.~33, pp. 17--34, 2024.

\bibitem{abdel2024matched}
M.~Abdel-Aty and S.~Ding, ``A matched case-control analysis of autonomous vs human-driven vehicle accidents,'' \emph{Nat. Commun.}, vol.~15, no.~1, p. 4931, 2024.

\bibitem{8370800}
S.~Noh, ``Decision-making framework for autonomous driving at road intersections: Safeguarding against collision, overly conservative behavior, and violation vehicles,'' \emph{IEEE Trans. Ind. Electron.}, vol.~66, no.~4, pp. 3275--3286, 2019.

\bibitem{20214811249569}
M.~T. Ashraf, K.~Dey, S.~Mishra, and M.~T. Rahman, ``Extracting rules from autonomousvehicle-involved crashes by applying decision tree and association rule methods,'' \emph{Transp. Res. Rec.}, vol. 2675, no.~11, pp. 522 -- 533, 2021.

\bibitem{10422331}
Z.~Li, L.~Xiong, B.~Leng, P.~Xu, and Z.~Fu, ``Safe reinforcement learning of lane change decision making with risk-fused constraint,'' in \emph{Proc. IEEE Intell. Transp. Syst. Conf.}, 2023, pp. 1313--1319.

\bibitem{20233614676836}
G.~Li, W.~Zhou, S.~Lin, S.~Li, and X.~Qu, ``On-ramp merging for highway autonomous driving: An application of a new safety indicator in deep reinforcement learning,'' \emph{Automot. Innov.}, vol.~6, no.~3, pp. 453 -- 465, 2023.

\bibitem{20244717409481}
H.~Hu, D.~Chu, J.~Yin, and L.~Lu, ``Double deep q-networks based game-theoretic equilibrium control of automated vehicles at autonomous intersection,'' \emph{Automot. Innov.}, vol.~7, no.~4, pp. 571 -- 587, 2024.

\bibitem{10740674}
K.~Yang, S.~Li, Y.~Chen, D.~Cao, and X.~Tang, ``Towards safe decision-making for autonomous vehicles at unsignalized intersections,'' \emph{IEEE Trans. Veh. Technol.}, vol.~74, no.~3, pp. 3830--3842, 2025.

\bibitem{10675394}
S.~Gu, L.~Yang, Y.~Du, G.~Chen, F.~Walter, J.~Wang, and A.~Knoll, ``A review of safe reinforcement learning: Methods, theories, and applications,'' \emph{IEEE Trans. Pattern Anal. Mach. Intell.}, vol.~46, no.~12, pp. 11\,216--11\,235, 2024.

\bibitem{stooke2020responsive}
A.~Stooke, J.~Achiam, and P.~Abbeel, ``Responsive safety in reinforcement learning by pid lagrangian methods,'' in \emph{Int. Conf. Mach. Learn.}\hskip 1em plus 0.5em minus 0.4em\relax PMLR, 2020, pp. 9133--9143.

\bibitem{honari2024meta}
H.~Honari, A.~M.~S. Enayati, M.~G. Tamizi, and H.~Najjaran, ``Meta sac-lag: Towards deployable safe reinforcement learning via metagradient-based hyperparameter tuning,'' in \emph{Proc. IEEE/RSJ Int. Conf. Intell. Robots Syst.}\hskip 1em plus 0.5em minus 0.4em\relax IEEE, 2024, pp. 619--626.

\bibitem{achiam2017constrained}
J.~Achiam, D.~Held, A.~Tamar, and P.~Abbeel, ``Constrained policy optimization,'' in \emph{Int. Conf. Mach. Learn.}\hskip 1em plus 0.5em minus 0.4em\relax PMLR, 2017, pp. 22--31.

\bibitem{zhang2020first}
Y.~Zhang, Q.~Vuong, and K.~Ross, ``First order constrained optimization in policy space,'' \emph{Adv. Neural Inf. Process. Syst.}, vol.~33, pp. 15\,338--15\,349, 2020.

\bibitem{shixin2024unmanned}
Z.~Shixin, P.~Feng, J.~Anni, Z.~Hao, and G.~Qiuqi, ``The unmanned vehicle on-ramp merging model based on am-mappo algorithm,'' \emph{Sci. Rep.}, vol.~14, no.~1, p. 19416, 2024.

\bibitem{10402567}
H.~Ma, C.~Liu, S.~E. Li, S.~Zheng, W.~Sun, and J.~Chen, ``Learn zero-constraint-violation safe policy in model-free constrained reinforcement learning,'' \emph{IEEE Trans. Neural Netw. Learn, Syst.}, vol.~36, no.~2, pp. 2327--2341, 2025.

\bibitem{10610959}
H.~Zheng, H.~Ma, S.~Zheng, S.~E. Li, and J.~Wang, ``Synthesize efficient safety certificates for learning-based safe control using magnitude regularization,'' in \emph{IEEE Int. Conf. Robot. Autom.}, 2024, pp. 545--551.

\bibitem{9718195}
X.~Wang, ``Ensuring safety of learning-based motion planners using control barrier functions,'' \emph{IEEE Robot. Autom. Lett.}, vol.~7, no.~2, pp. 4773--4780, 2022.

\bibitem{8796030}
A.~D. Ames, S.~Coogan, M.~Egerstedt, G.~Notomista, K.~Sreenath, and P.~Tabuada, ``Control barrier functions: Theory and applications,'' in \emph{Eur. Control Conf. (ECC)}, 2019, pp. 3420--3431.

\bibitem{10104197}
W.~Zhou, Z.~Cao, N.~Deng, K.~Jiang, and D.~Yang, ``Identify, estimate and bound the uncertainty of reinforcement learning for autonomous driving,'' \emph{IEEE Trans. Intell. Transport. Syst.}, vol.~24, no.~8, pp. 7932--7942, 2023.

\bibitem{hullermeier2021aleatoric}
E.~H{\"u}llermeier and W.~Waegeman, ``Aleatoric and epistemic uncertainty in machine learning: An introduction to concepts and methods,'' \emph{Mach. Learn.}, vol. 110, no.~3, pp. 457--506, 2021.

\bibitem{10836828}
S.~Wang and S.~Wen, ``Safe control against uncertainty: A comprehensive review of control barrier function strategies,'' \emph{IEEE Syst. Man Cybern. Mag.}, vol.~11, no.~1, pp. 34--47, 2025.

\bibitem{10073955}
C.-J. Hoel, K.~Wolff, and L.~Laine, ``Ensemble quantile networks: Uncertainty-aware reinforcement learning with applications in autonomous driving,'' \emph{IEEE Trans. Intell. Transport. Syst.}, vol.~24, no.~6, pp. 6030--6041, 2023.

\bibitem{10534899}
Z.~Zhang, Q.~Liu, Y.~Li, K.~Lin, and L.~Li, ``Safe reinforcement learning in autonomous driving with epistemic uncertainty estimation,'' \emph{IEEE Trans. Intell. Transport. Syst.}, vol.~25, no.~10, pp. 13\,653--13\,666, 2024.

\bibitem{10107652}
K.~Yang, X.~Tang, S.~Qiu, S.~Jin, Z.~Wei, and H.~Wang, ``Towards robust decision-making for autonomous driving on highway,'' \emph{IEEE Trans. Veh. Technol.}, vol.~72, no.~9, pp. 11\,251--11\,263, 2023.

\bibitem{10155311}
X.~Tang, G.~Zhong, S.~Li, K.~Yang, K.~Shu, D.~Cao, and X.~Lin, ``Uncertainty-aware decision-making for autonomous driving at uncontrolled intersections,'' \emph{IEEE Trans. Intell. Transport. Syst.}, vol.~24, no.~9, pp. 9725--9735, 2023.

\bibitem{ganaie2022ensemble}
M.~A. Ganaie, M.~Hu, A.~K. Malik, M.~Tanveer, and P.~N. Suganthan, ``Ensemble deep learning: A review,'' \emph{Eng. Appl. Artif. Intell.}, vol. 115, p. 105151, 2022.

\bibitem{ma2020dsac}
X.~Ma, L.~Xia, Z.~Zhou, J.~Yang, and Q.~Zhao, ``Dsac: Distributional soft actor critic for risk-sensitive reinforcement learning,'' \emph{arXiv preprint arXiv:2004.14547}, 2020.

\bibitem{huber1992robust}
P.~J. Huber, ``Robust estimation of a location parameter,'' in \emph{Breakthroughs in statistics: Methodology and distribution}.\hskip 1em plus 0.5em minus 0.4em\relax Springer, 1992, pp. 492--518.

\bibitem{9516971}
W.~Xiao and C.~Belta, ``High-order control barrier functions,'' \emph{IEEE Trans. Autom. Control.}, vol.~67, no.~7, pp. 3655--3662, 2022.

\bibitem{9777251}
Y.~Xiong, D.-H. Zhai, M.~Tavakoli, and Y.~Xia, ``Discrete-time control barrier function: High-order case and adaptive case,'' \emph{IEEE Trans. Cybern.}, vol.~53, no.~5, pp. 3231--3239, 2023.

\bibitem{ROCKAFELLAR20021443}
R.~Rockafellar and S.~Uryasev, ``Conditional value-at-risk for general loss distributions,'' \emph{J. Bank. Financ.}, vol.~26, no.~7, pp. 1443--1471, 2002.

\bibitem{rame2021dice}
A.~Rame and M.~Cord, ``Dice: Diversity in deep ensembles via conditional redundancy adversarial estimation,'' in \emph{Proc. Int. Conf. Learn. Represent.}, 2021.

\bibitem{osband2016deep}
I.~Osband, C.~Blundell, A.~Pritzel, and B.~Van~Roy, ``Deep exploration via bootstrapped dqn,'' \emph{Adv. Neural Inf. Process. Syst.}, vol.~29, 2016.

\bibitem{osband2018randomized}
I.~Osband, J.~Aslanides, and A.~Cassirer, ``Randomized prior functions for deep reinforcement learning,'' \emph{Adv. Neural Inf. Process. Syst.}, vol.~31, 2018.

\bibitem{rajamani2011vehicle}
R.~Rajamani, \emph{Vehicle dynamics and control}.\hskip 1em plus 0.5em minus 0.4em\relax Springer Science \& Business Media, 2011.

\bibitem{xu2025high}
J.~Xu and B.~Alrifaee, ``High-order control barrier functions: Insights and a truncated taylor-based formulation,'' \emph{arXiv preprint arXiv:2503.15014}, 2025.

\bibitem{Andersson2018}
J.~A.~E. Andersson, J.~Gillis, G.~Horn, J.~B. Rawlings, and M.~Diehl, ``{CasADi} -- {A} software framework for nonlinear optimization and optimal control,'' \emph{Mathematical Programming Computation}, 2018.

\bibitem{highwayenv}
E.~Leurent, ``An environment for autonomous driving decision-making,'' \url{https://github.com/eleurent/highway-env}, 2018.

\bibitem{idm}
M.~Treiber, A.~Hennecke, and D.~Helbing, ``Congested traffic states in empirical observations and microscopic simulations,'' \emph{Phys. Rev. E}, pp. 1805--1824, Jul 2002.

\bibitem{leng2025risk}
B.~Leng, R.~Yu, W.~Han, L.~Xiong, Z.~Li, and H.~Huang, ``Risk-aware reinforcement learning for autonomous driving: Improving safety when driving through intersection,'' \emph{arXiv preprint arXiv:2503.19690}, 2025.

\end{thebibliography}
\end{document}